%% file: main.tex
\documentclass[10pt,twocolumn,letterpaper]{article}

\usepackage[pagenumbers]{cvpr}
\usepackage{graphicx}
\usepackage{booktabs}
\usepackage{multirow}
\usepackage[table]{xcolor}
\definecolor{gainGreen}{HTML}{16803C}

\input{preamble}
\definecolor{cvprblue}{rgb}{0.21,0.49,0.74}
\usepackage[pagebackref,breaklinks,colorlinks,allcolors=cvprblue]{hyperref}

\def\confName{CVPR}
\def\confYear{2027}

\title{ProSG: Learning Progressive Structural Guidance for 3D Human Pose Estimation}

\author{
Zehua Wang$^{1}$ \quad
Qiongjie Cui$^{2}$ \quad
Changwang Mei$^{1}$ \quad
Pengqi Hu$^{1}$ \quad
Huaijiang Sun$^{1}$\\
$^{1}$Nanjing University of Science and Technology\\
$^{2}$Xi'an Jiaotong University\\
{\ttfamily\footnotesize
relaxwang0714@gmail.com \quad
cuiqiongjie@xjtu.edu.cn}\\
{\ttfamily\footnotesize
meichangwang@njust.edu.cn \quad
hupengqi@njust.edu.cn \quad
sunhuaijiang\_njust@163.com}
}

\begin{document}
\maketitle
\input{sec/0_abstract}    
\input{sec/1_intro}

\input{sec/2_related_work}

\input{sec/3_method_old}
\input{sec/4_experiments}

\input{sec/conclusion}


{
    \small
    \bibliographystyle{ieeenat_fullname}
    \bibliography{main}
}


\end{document}

%% file: sec/0_abstract.tex
\begin{abstract}
For 3D human pose estimation from detected 2D joint locations, depth ambiguity remains a fundamental challenge. Even when two connected joints are accurately localized in the image, their relative depth along the camera axis can remain uncertain. Existing methods progressively refine pose representations through successive stages, with early stages emphasizing local 2D skeletal geometry and joint relationships and deeper stages developing increasingly depth-related representations. However, early-stage structural cues remain overlooked despite their importance for subsequent 3D inference. Our findings reveal that reusing such information provides complementary geometric guidance, particularly improving joint position estimation along the camera-space depth axis. We propose ProSG, a progressive structural guidance framework that preserves and aggregates intermediate pose representations from preceding stages to guide subsequent 3D pose refinement.  Specifically, Structure Guided Accumulation (SGA) maintains a pool of these representations and adaptively retrieves relevant information for each joint and frame, enabling deeper stages to revisit early structural cues.  Structural Context Reorganization (SCR) further reorganizes intermediate representations before they enter the representation pool, using compact structural contextual tokens to facilitate the effective integration of structural information across stages. Extensive experiments on Human3.6M and MPI-INF-3DHP demonstrate the effectiveness of ProSG, achieving state-of-the-art performance.


\end{abstract}

\begin{figure}[ht!]
  \centering
   \includegraphics[width=\linewidth]{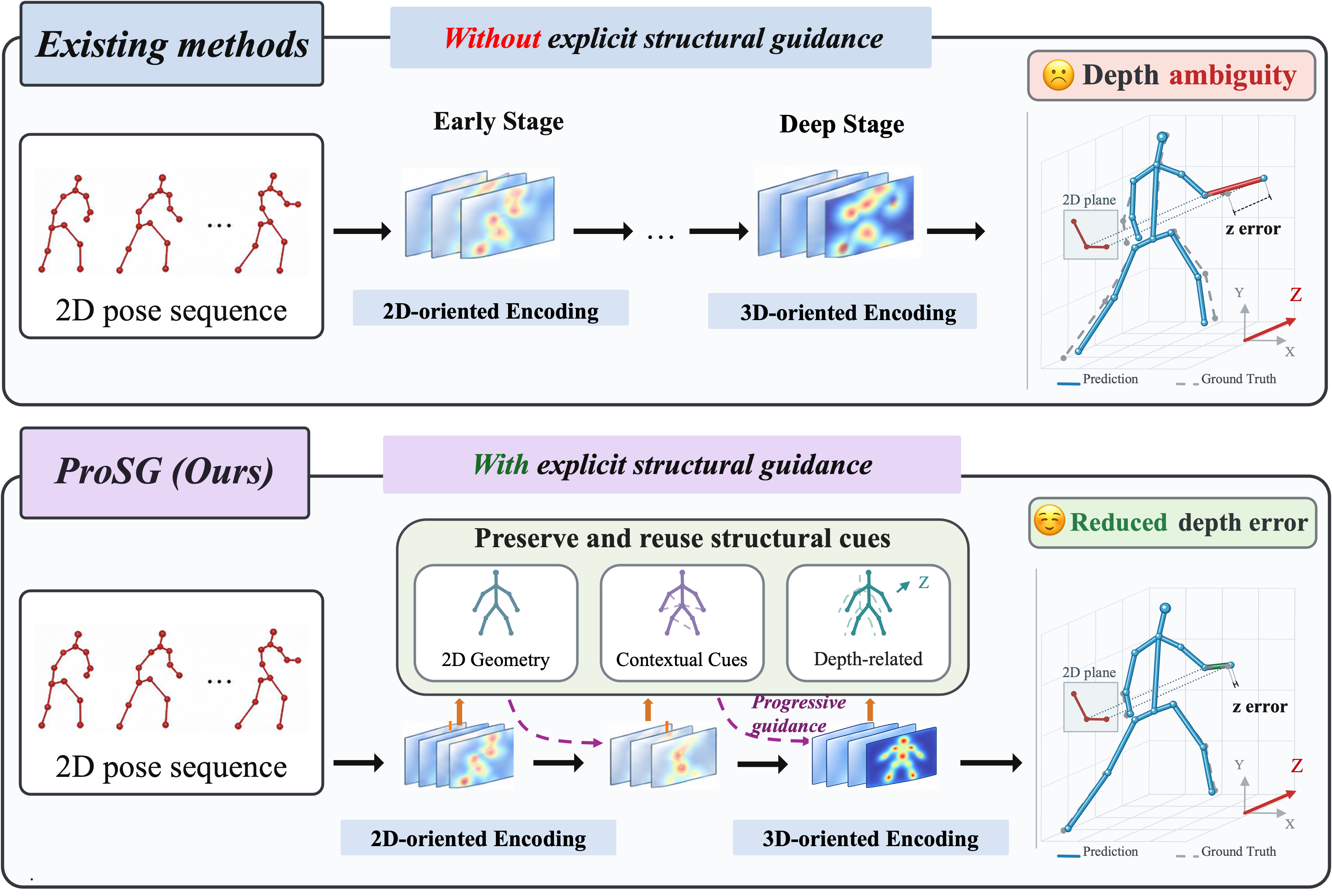}
\vspace{-1.5em}

\caption{
(Top) Existing methods refine pose representations
without explicitly leveraging structural cues for
subsequent 3D inference.
(Bottom) ProSG integrates structural cues across stages to improve
3D pose estimation, particularly along the camera-space
depth axis.}

   \label{fig:motivation}
   \vspace{-1.5em}
\end{figure}

%% file: sec/1_intro.tex
\section{Introduction}
\label{sec:intro}
3D human pose estimation aims to recover the spatial configuration of human joints and plays an important role in applications such as human–computer interaction~\cite{chen2019so,ye2021collaborative,chen2021joint}, motion analysis~\cite{yeung2024autosoccerpose,yeung2025athletepose3d}, and virtual reality~\cite{zhang2022unsupervised,zhang2022spatial,jiang2022avatarposer,zhang2022egobody}. 2D-to-3D pose lifting has emerged as a paradigm that reconstructs 3D poses from detected 2D joint sequences.

A fundamental challenge lies in depth ambiguity~\cite{liu2025posemoe,shan2023diffusion,yang2024kitro,liao2026langpose}: while 2D observations specify joint locations in the image plane, their camera-space depths remain unknown. 
n particular, the relative depth of connected joints may remain ambiguous despite their observed 2D positions, potentially leading to erroneous out-of-plane limb orientations.
Therefore, effectively exploiting the structural information embedded in 2D observations, including local geometry and joint connectivity, is crucial for inferring joint depth and reconstructing 3D poses.
\begin{figure*}[t]
    \centering
    \includegraphics[width=\textwidth]{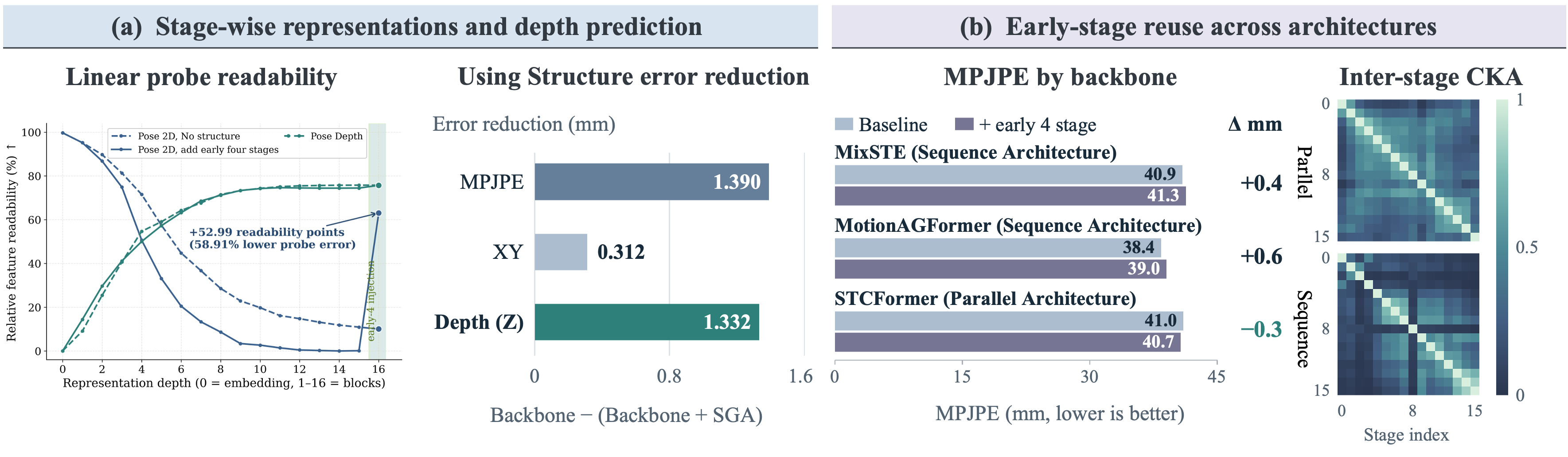}
    \vspace{-1.5em}
    \vspace{-1em}
\caption{\textbf{Motivation for progressive structural guidance.}
(a) Linear probing indicates that 2D geometric information is more readily decodable in early stages. Probing details are provided in the supplementary material.
(b) Reusing preceding representations yields architecture-dependent gains associated with inter-stage similarity measured by Centered Kernel Alignment(CKA).}

    \label{fig:introduction}
    \vspace{-1em}
\end{figure*}

Existing approaches have incorporated structural relationships into pose representation learning \cite{li2025graphmlp,chen2019so,peng2024ktpformer,jang2025remp,wang20243d,zhao2022graformer,zheng2025hipart,zhai2023hopfir}. MotionAGFormer combines Transformer and graph-based modules to capture global and local joint interactions \cite{mehraban2024motionagformer}, while KTPFormer incorporates kinematic and trajectory priors into its attention mechanisms \cite{peng2024ktpformer}. While these designs effectively encode structural joint relationships, a fundamental question remains: how is this structural information utilized as features evolve across successive processing stages? 
As shown in Fig.~\ref{fig:motivation}, existing methods relying on standard residual propagation lack direct, selective access to early structural cues. Successive residual updates mix these cues into accumulated representations, limiting their availability for constraining relative joint depth.  This may lead to incorrect relative depth estimates for joints connected by bones nearly parallel to the image plane.


To investigate this question, we analyze pose representations across successive network stages. Our stage-wise linear probes\cite{alain2016understanding,zhu2024revisiting} reveal that 2D geometric information is more readily decodable in early stages, whereas depth-related information becomes increasingly decodable in deeper representations, as shown in Fig.~\ref{fig:introduction}(a). Moreover, reusing early-stage representations improves the decodability of 2D geometric information in the final representation and yields a more pronounced reduction in prediction errors along the camera-space depth axis than along the image-plane axes. These findings suggest that early-stage representations retain complementary geometric cues that can provide additional guidance for camera-space depth estimation. More broadly, as pose representations evolve through successive stages, different intermediate representations may contain complementary structural information, motivating its explicit reuse across network stages.

However, effectively integrating structural information embedded in intermediate representations requires more than simply preserving and retrieving preceding features. As pose features undergo successive spatio-temporal transformations, their contextual organization may vary, potentially affecting the effectiveness of cross-stage aggregation. To examine this issue, we investigate 	structural information reuse across different spatio-temporal architectures. As shown in Fig.~\ref{fig:introduction}(b), the improvements vary across architectures, with larger gains associated with higher inter-stage representation similarity measured by CKA \cite{kornblith2019similarity,jiang2025tracing}. These observations motivate a framework that not only selectively retrieves complementary structural information from preceding stages, but also organizes intermediate features to facilitate their integration into subsequent representations.

We propose ProSG, a progressive structural guidance framework that leverages complementary pose representations across network stages to improve 3D human pose estimation. Considering the observed relationship between inter-stage representation similarity and aggregation effectiveness, we adopt a parallel spatio-temporal Transformer backbone to facilitate structural information integration. Specifically, Structure Guided Accumulation (SGA) maintains a pool of intermediate pose representations and adaptively aggregates them across successive stages. By learning structure-aware weights for individual joint-frame tokens, SGA selectively integrates complementary structural cues from preceding representations to guide subsequent pose refinement. Unlike Attention Residual~\cite{team2026attention}, which operates at the attention and MLP sublayer level, SGA aggregates representations from complete spatio-temporal blocks while preserving joint-frame correspondence. To facilitate effective structural guidance, we introduce Structural Context Reorganization (SCR), which explicitly organizes the structural information embedded in intermediate pose representations before they enter SGA. This process reorganizes the encoded spatio-temporal context while preserving joint-frame correspondence, enabling more compatible integration of intermediate representations.

Our contributions are summarized as follows: \textbf{1. }We investigate the evolution of pose representations across network stages, identifying complementary 2D geometric and depth-related information and highlighting the potential of progressively reusing earlier structural cues to guide subsequent 3D inference. \textbf{2. }We propose ProSG for progressive structural guidance, featuring joint-frame-level adaptive accumulation of structural information (SGA) and structured intermediate representation organization (SCR). \textbf{3. }Extensive experiments on Human3.6M and MPI-INF-3DHP demonstrate  ProSG achieving state-of-the-art performance with a favorable accuracy-efficiency trade-off.






%% file: sec/2_related_work.tex
\section{Related Work}
\label{sec:related work}
\textbf{2D-to-3D Pose Lifting} Existing 3D human pose estimation methods can be broadly categorized into one-stage and two-stage approaches. One-stage methods directly regress 3D joint coordinates from images or videos~\cite{lin2021end,sun2018integral,yoshiyasu2023deformable,kanazawa2018end,sun2024aios,dwivedi2024tokenhmr,Hao_2025_ICCV,Xia_2025_CVPR}. In contrast, two-stage methods first detect 2D joint locations using pose estimators such as CPN~\cite{chen2018cascaded} and SH~\cite{newell2016stacked}, and subsequently lift them into 3D space. Despite their effectiveness, recovering 3D poses from 2D observations remains inherently ambiguous due to the loss of depth information. To alleviate this ambiguity, Transformer-based methods~\cite{zheng20213d,zhang2022mixste,zheng2025spectral,chen2023hdformer,zhao2023poseformerv2,zhang2024app,li2022exploiting,tang2023ftcm,pavllo20193d,li2024hourglass} exploit spatio-temporal dependencies, while recent state space models (SSMs)~\cite{lu2025structure,zhang2025pose,huang2025posemamba,cui2025hgmamba,cui2026sasmamba} offer efficient sequence modeling. Our work follows the two-stage paradigm and investigates progressive structural guidance for 3D pose reconstruction.

\textbf{Transformer-based 3D Pose Estimation.}
Transformer-based methods model spatial and temporal dependencies through different architectures. PoseFormer~\cite{zheng20213d} employs spatial and temporal Transformers, MixSTE~\cite{zhang2022mixste} alternates spatial and temporal modeling, and STCFormer~\cite{tang20233d} processes both dependencies in parallel. MotionBERT~\cite{zhu2023motionbert} and SCTFormer~\cite{zheng2025spectral} explore complementary representation streams. Other methods incorporate structural priors or specialized interactions. MotionAGFormer~\cite{mehraban2024motionagformer} combines Transformer and GCN branches, KTPFormer~\cite{peng2024ktpformer} introduces kinematic and trajectory priors, and TCPFormer~\cite{liu2025tcpformer} uses implicit pose proxies for temporal interaction. These methods primarily investigate how joint relationships and motion context are encoded during feature transformation. Our work addresses a complementary aspect: how intermediate structural representations can be selectively reused across reconstruction stages. ProSG aggregates these representations at corresponding joint-frame positions, with skeletal-neighbor scoring and temporal smoothing refining their contributions. This cross-stage guidance complements structural modeling within individual stages.

\textbf{Feature Aggregation and Context Modeling.}
Attention Residual~\cite{team2026attention} replaces fixed residual
updates with adaptive aggregation. Its aggregation is designed
to combine features produced by the attention and MLP sublayers.
In contrast, ProSG aggregates complete pose-stage representations,
using skeletal and temporal context to determine joint-frame-specific weights. Unlike other context modeling
methods~\cite{lee2019settransformer,jaegle2021perceiver},
SCR derives representatives from current-stage queries through
token-axis projection rather than learned initial vectors.
For pose estimation, its spatial and temporal branches reorganize
joint-configuration and trajectory context through gathering and
redistribution, preserving joint-frame correspondence to support
structural guidance across reconstruction stages.

%% file: sec/3_method_old.tex
\section{Proposed Method}

\begin{figure*}[h]
  \centering
  \vspace{-1em}
   \includegraphics[width=\linewidth]{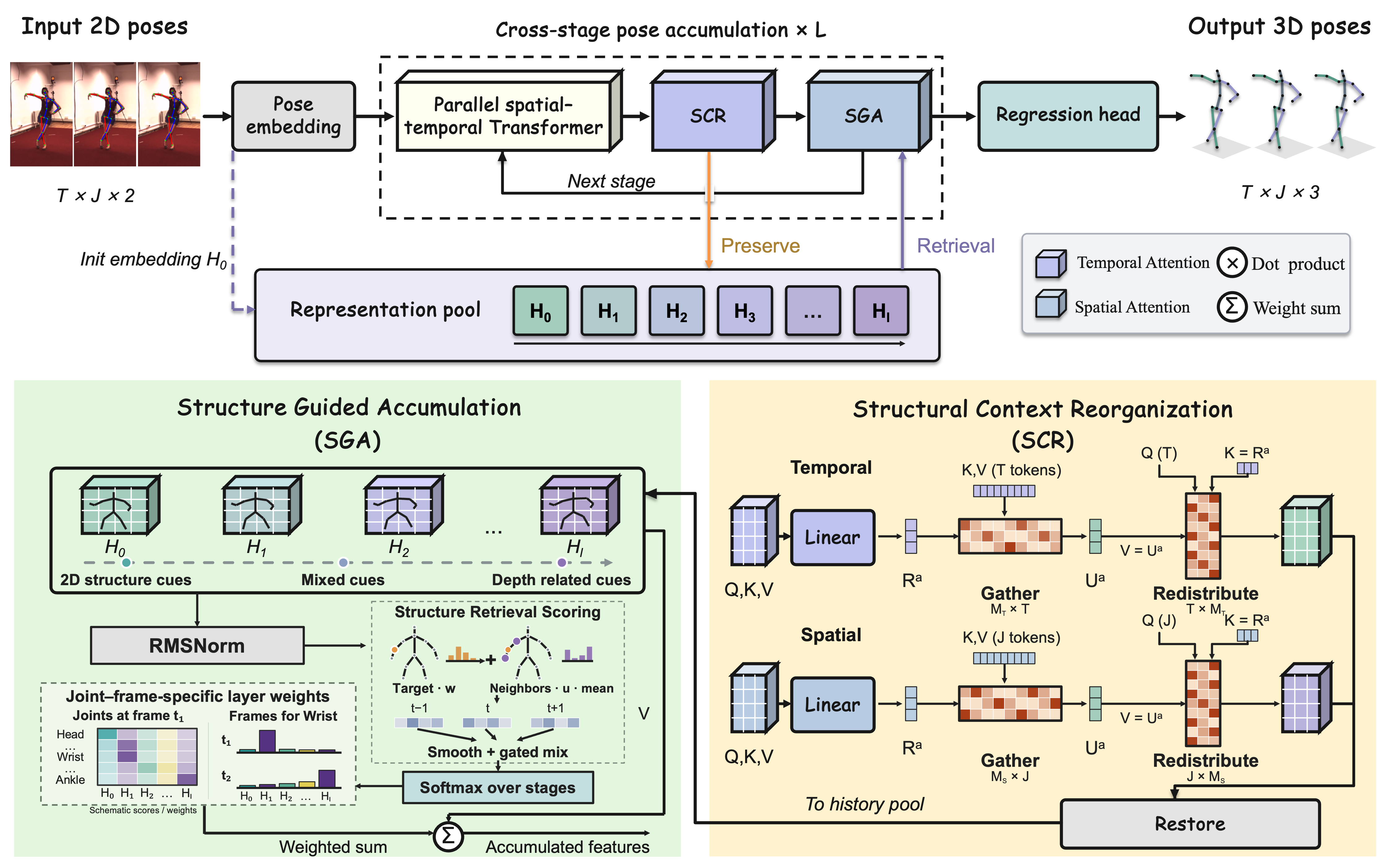}
\vspace{-2em}
   \caption{Overview of the proposed ProSG framework. Each stage retains its first spatio-temporal (ST) block and replaces the second with Structural Context Reorganization (SCR), whose output is stored in a representation pool. Structure Guided
Accumulation (SGA) selectively aggregates the stored
representations at corresponding joint-frame positions to
guide subsequent stages of 3D pose reconstruction. }
   \label{fig:onecol}
   \vspace{-1em}
\end{figure*}

\subsection{Overview}
\label{sec:framework_overview}

Given a sequence of 2D poses, ProSG predicts the corresponding
3D joint coordinates through progressive structural guidance.
The input is embedded into pose features and retained as the
initial representation. SCR replaces the last block at each stage in the parallel backbone to reorganize structural information. At each stage, SGA aggregates the available representations to construct the next-stage input. The final features are mapped to 3D coordinates by a regression head. Fig.~\ref{fig:onecol} illustrates the overall computation.

\textbf{Parallel Spatio-Temporal Backbone.} Motivated by the observed relationship between inter-stage representation similarity and aggregation effectiveness, our backbone adopts the parallel spatial and temporal grouping strategy of STCFormer~\cite{tang20233d}. Given an input sequence of 2D poses, the embedded features are represented as $\mathbf{X}\in\mathbb{R}^{(T\times J)\times C}$, where $T$, $J$, and $C$ denote the numbers of frames, joints, and feature channels, respectively. The backbone consists of $L$ stacked stages, with attention heads divided into spatial and temporal groups to model joint dependencies and motion dynamics in parallel.


Although both our backbone and STCFormer employ depthwise convolution, their effective receptive fields differ.
STCFormer applies a $3\times3$ depthwise convolution
to each half-channel value tensor of shape
$B\times(C/2)\times T\times J$.
Consequently, the output at position $(t,j)$ combines
features from neighboring frames and neighboring
joint indices in both branches.

In contrast, we reshape the spatial and temporal
value features as
\begin{equation}
\begin{aligned}
\mathbf{V}^{s}
&\in\mathbb{R}^{(BT)\times(C/2)\times1\times J},\\
\mathbf{V}^{t}
&\in\mathbb{R}^{(BJ)\times(C/2)\times T\times1}.
\end{aligned}
\end{equation}
With these layouts, the same $3\times3$ convolution
has an effective receptive field of $1\times3$
in the spatial branch and $3\times1$ in the temporal
branch.
The former aggregates neighboring joint indices
only within the same frame, whereas the latter
aggregates neighboring frames only for the same joint.
The branch outputs are computed as
\begin{equation}
\begin{aligned}
\mathbf{O}^{s}
&=
\mathcal{S}
(\mathbf{Q}^{s},\mathbf{K}^{s},\mathbf{V}^{s})
+
\mathcal{D}_{s}(\mathbf{V}^{s}),\\
\mathbf{O}^{t}
&=
\mathcal{T}
(\mathbf{Q}^{t},\mathbf{K}^{t},\mathbf{V}^{t})
+
\mathcal{D}_{t}(\mathbf{V}^{t}),
\end{aligned}
\label{eq:axis_decoupled_modeling}
\end{equation}
where $\mathcal{S}$ and $\mathcal{T}$ denote spatial
and temporal multi-head self-attention, respectively,
returning concatenated head outputs before the shared
output projection.
$\mathcal{D}_{s}$ and $\mathcal{D}_{t}$ denote the
corresponding axis-decoupled depthwise convolutions.


\subsection{Structure Guided Accumulation}
\label{sec:SGA}

Structure Guided Accumulation (SGA) provides subsequent
reconstruction stages with direct access to structure-bearing representations preserved from preceding stages.
These representations retain explicit joint-frame organization,
while skeletal relationships and temporal context further
refine their contributions during structural accumulation. Let $\mathcal{H}_{\ell}
=[\mathbf{H}_0,\ldots,\mathbf{H}_{\ell}]$
denote the representation pool after stage $\ell$,
where $\mathbf{H}_0$ is the initial pose embedding
and $\mathbf{H}_i\in\mathbb{R}^{B\times T\times J\times C}$
is the completed stage-wise structural representation organized
by SCR before entering the pool. 

\textbf{Structure Retrieval Scoring}
For each joint-frame position $(t,j)$, historical
structural information is evaluated using both the
target joint and its directly connected skeletal neighbors.
Let
$\mathbf{K}_{\ell,i}^{(t,j)}
=\operatorname{RMSNorm}_{\ell}
(\mathbf{H}_{i}^{(t,j)})$
denote channel-normalized features~\cite{zhang2019root}.
The score for stage-wise structural representations $i$ is
\begin{equation}
\begin{aligned}
s_{\ell,i}^{(t,j)}
={}&
\mathbf{w}_{\ell}^{\top}
\mathbf{K}_{\ell,i}^{(t,j)}
+
\frac{1}{|\mathcal{N}(j)|}
\sum_{k\in\mathcal{N}(j)}
\mathbf{u}_{\ell}^{\top}
\mathbf{K}_{\ell,i}^{(t,k)},
\end{aligned}
\label{eq:sga_scoring}
\end{equation}
where $\mathcal{N}(j)$ contains the one-hop skeletal
neighbors of joint $j$, excluding itself.
The learnable vectors
$\mathbf{w}_{\ell},\mathbf{u}_{\ell}\in\mathbb{R}^{C}$
separately score target-joint and neighbor features.
Neighbor scores are averaged to account for different
joint degrees.
The neighbor term allows connected-joint context at
each historical stage to influence how strongly that
stage contributes to the target joint's representation.

We further incorporate local temporal context by
smoothing these structure-informed scores across
adjacent frames, where $\gamma \in [0, \tfrac12]$ controls the contribution
of neighboring frames and is set to $0.25$ in our experiments:
\begin{equation}
\bar{s}_{\ell,i}^{(t,j)}
=
\gamma s_{\ell,i}^{(t-1,j)}
+
(1-2\gamma)s_{\ell,i}^{(t,j)}
+
\gamma s_{\ell,i}^{(t+1,j)},
\label{eq:sga_smoothing}
\end{equation}
The original and smoothed scores are combined as
\begin{equation}
\widetilde{s}_{\ell,i}^{(t,j)}
=
(1-\lambda_{\ell})s_{\ell,i}^{(t,j)}
+
\lambda_{\ell}\bar{s}_{\ell,i}^{(t,j)},
\label{eq:sga_mixing}
\end{equation}
where $\lambda_{\ell}=\sigma(\beta_{\ell})$
is a learnable scalar controlling the smoothing strength.
Boundary scores are replicated at sequence endpoints.
We initialize $\beta_{\ell}$ to $-2$ and both scoring
vectors to zero, yielding a small initial smoothing
contribution and uniform structure weights.

\textbf{Structural Information Aggregation.}
The resulting scores are normalized across previous
stages to construct the next-stage input:
\begin{equation}
\begin{aligned}
\alpha_{\ell,i}^{(t,j)}
&=
\frac{
\exp(\widetilde{s}_{\ell,i}^{(t,j)})
}{
\sum_{k=0}^{\ell}
\exp(\widetilde{s}_{\ell,k}^{(t,j)})
},\\
\mathbf{X}_{\ell}^{(t,j)}
&=
\sum_{i=0}^{\ell}
\alpha_{\ell,i}^{(t,j)}
\mathbf{H}_{i}^{(t,j)},
\qquad \ell<L.
\end{aligned}
\label{eq:sga_aggregation}
\end{equation}
This formulation accumulates structural representations at corresponding joint-frame positions. Skeletal and temporal context further refine their relative contributions without disrupting the target joint's correspondence across stages.

While AttnRes~\cite{team2026attention} focuses on adaptive
aggregation of sublayer features, SGA aggregates and retrieves complete spatio-temporal stage representations reorganized by SCR
to guide pose reconstruction, with skeletal-neighbor scoring
and temporal smoothing refining each stage's contribution.

\subsection{Structural Context Reorganization}
\label{sec:SCR}


Although SGA adaptively aggregates intermediate pose representations at corresponding joint-frame positions, their encoded spatio-temporal context may differ across network stages, potentially limiting effective integration. We therefore introduce Structural Context Reorganization (SCR), which gathers and redistributes contextual information before intermediate representations enter the representation pool. This process enriches their structural context while preserving joint-frame correspondence, facilitating more compatible 	cross-stage structural guidance.

For each branch $a\in\{s,t\}$, we project the output features
of first ST block in the backbone into
$\mathbf{Q}^a,\mathbf{K}^a,\mathbf{V}^a
\in\mathbb{R}^{N_a\times d}$,
where $N_s=J$, $N_t=T$, and $d$ is the dimension per head.
Contextual representatives are constructed as
\begin{equation}
\mathbf{R}^{a}
=
\mathbf{W}_{p}^{a}\mathbf{Q}^{a}
\in\mathbb{R}^{M_a\times d},
\label{eq:scr_representatives}
\end{equation}
where $\mathbf{W}_{p}^{a}\in\mathbb{R}^{M_a\times N_a}$
maps branch positions to $M_a$ representatives. We define two contextual mappings:
\begin{equation}
\begin{aligned}
\mathbf{G}^{a}
&=
\rho_{N_a}
\left(
\frac{\mathbf{R}^{a}(\mathbf{K}^{a})^\top}{\sqrt{d}}
+\mathbf{B}_{g}^{a}
\right),\\
\mathbf{D}^{a}
&=
\rho_{M_a}
\left(
\frac{\mathbf{Q}^{a}(\mathbf{R}^{a})^\top}{\sqrt{d}}
+\mathbf{B}_{r}^{a}
\right),
\end{aligned}
\label{eq:scr_mappings}
\end{equation}
where $\rho_n$ denotes row-wise softmax normalization
over $n$ positions, and
$\mathbf{B}_{g}^{a}$ and $\mathbf{B}_{r}^{a}$
encode spatial and temporal relational priors through learnable biases.

\noindent\textbf{Gathering.}
The gathering map
$\mathbf{G}^{a}\in\mathbb{R}^{M_a\times N_a}$
collects pose information into contextual representatives:
\begin{equation}
\mathbf{U}^{a}=\mathbf{G}^{a}\mathbf{V}^{a}.
\label{eq:scr_gather}
\end{equation}
These representatives summarize joint-configuration
context within each frame or trajectory context
for each joint.

\noindent\textbf{Redistribution.}
The redistribution map
$\mathbf{D}^{a}\in\mathbb{R}^{N_a\times M_a}$
returns the gathered context to the original positions:
\begin{equation}
\mathbf{Z}^{a}
=
\mathbf{D}^{a}\mathbf{U}^{a}
=
\underbrace{\mathbf{D}^{a}\mathbf{G}^{a}}_{\mathbf{M}^{a}}
\mathbf{V}^{a}.
\label{eq:scr_redistribute}
\end{equation}
The effective mapping $\mathbf{M}^{a}$ enriches
each position with representative-mediated context
while preserving joint-frame correspondence.
The branch outputs are fused to form the refined stage representation stored in the pool.

\textbf{Regression Head and Loss Function}
After feature encoding, a linear regression head is then applied to predict the 3D pose sequence. For training, we adopt the same loss formulation as MotionAGFormer~\cite{mehraban2024motionagformer}. Specifically, we minimize the $\ell_2$ distance between the predicted and ground-truth 3D joint coordinates, and further incorporate a velocity-based loss to enforce temporal consistency across frames.

%% file: sec/4_experiments.tex
\section{Experiments}
\subsection{Datasets and Evaluation Metrics}


\textbf{Human3.6M.} Human3.6M~\cite{ionescu2013human3} is the most established indoor benchmark, containing 3.6 million 3D human poses of 11 subjects performing 15 daily activities, captured by four synchronized cameras. Following standard protocols~\cite{zheng20213d,huang2025posemamba,li2022mhformer,liu2025tcpformer}, we use subjects S1, S5, S6, S7, and S8 for training, and evaluate on subjects S9 and S11. We report two widely used metrics: (P1) Mean Per-Joint Position Error (MPJPE, mm, ↓), which measures the average Euclidean distance between predicted and ground-truth 3D joint positions after alignment, and (P2) Procrustes-aligned MPJPE (P-MPJPE, mm, ↓), which further aligns the predicted pose to the ground truth via a rigid transformation.

\textbf{MPI-INF-3DHP.} MPI-INF-3DHP~\cite{mehta2017monocular} is a large-scale dataset covering both indoor and outdoor scenes with more diverse and challenging settings. It contains over 1.3 million frames from multiple subjects performing various activities, captured by 14 cameras. Following prior works~\cite{zheng20213d,lu2025structure,zhang2022mixste,mehraban2024motionagformer,peng2024ktpformer,martinez2017simple}, we evaluate our method using MPJPE (mm, ↓), Percentage of Correct Keypoints (PCK, \%, ↑) under a 150 mm threshold, and the Area Under the Curve (AUC, \%, ↑) over varying PCK thresholds.

\begin{table*}[ht!]
\centering
\caption{Comparison with state-of-the-art methods on Human3.6M
under P1 (MPJPE).
Unmarked methods use SH detections;
$\ast$ and $\ddagger$ indicate CPN and HRNet detections,
respectively. P1$^\dagger$ reports MPJPE with
ground-truth 2D inputs.
MACs/frame represents the number of multiply--accumulate
operations per output frame.
The best and second-best accuracy results are highlighted in
\textbf{bold} and \underline{underlined}.}
\label{tab:computational-cost}

\small
\setlength{\tabcolsep}{4pt}
\renewcommand{\arraystretch}{1.1}

\resizebox{0.9\textwidth}{!}{
\begin{tabular}{l|l|c|c|c|c|c|c}
\toprule
Method & Year & T
& Params (M)$\downarrow$
& MACs (G)$\downarrow$
& MACs/frame (M)$\downarrow$
& P1 (mm)$\downarrow$
& P1$^{\dagger}$ (mm)$\downarrow$ \\
\midrule

MixSTE$^{*}$~\cite{zhang2022mixste}
& 2022 & 243 & 33.8 & 147.6 & 607 & 40.9
& 21.6 \\

MotionBERT-finetune~\cite{zhu2023motionbert}
& 2023 & 243 & 42.5 & 349.5 & 1436 & 37.5
& 16.9 \\

STCFormer$^{*}$~\cite{tang20233d}
& 2023 & 243 & 4.7 & 19.6 & 80 & 41.0
& 21.3 \\

GLA-GCN$^{*}$~\cite{yu2023gla}
& 2023 & 243 & 1.3 & 1.5 & 6 & 44.4
& 21.0 \\

MotionAGFormer-B~\cite{mehraban2024motionagformer}
& 2024 & 243 & 11.7 & 64.8 & 266 & 38.4
& 19.4 \\

KTPFormer$^{*}$~\cite{peng2024ktpformer}
& 2024 & 243 & 33.7 & 139.1 & 572 & 40.1
& 19.0 \\
TCPFormer~\cite{liu2025tcpformer}
& 2025 & 243 & 35.1 & 109.2 & 449 & 37.9
& 15.5 \\


PoseMamba-L~\cite{huang2025posemamba}
& 2025 & 243 & 27.9 & 11.5 & 47 & 38.1
& 15.6 \\


PoseMagic~\cite{zhang2025pose}
& 2025 & 243 & 14.4 & 40.5 & 166 & 37.5
& -- \\

SAMA-S~\cite{lu2025structure}
& 2025 & 243 & \textbf{1.1} & \textbf{3.9} & \textbf{16} & 40.6
& 20.2 \\

SAMA-L~\cite{lu2025structure}
& 2025 & 243 & 17.3 & 53.2 & 218 & 36.9
& \underline{11.9} \\

SCTFormer~\cite{zheng2025spectral}
& 2026 & 243 & 42.6 & 58.9 & 242 & 37.7
& 15.4 \\

MSTPFormer$^{\ddagger}$~\cite{peng2026mstpformer}
& 2026 & 243 & 15.7 & 57.1 & 235 & 37.6
& 13.4 \\

LDOFormer$^{*}$~\cite{zou2026ldoformer}
& 2026 & 243 & 3.7 & -- & 68 & 37.5
& -- \\

\midrule

\textbf{ProSG-S}
& -- & 81 & 1.7 & 2.0 & 24 & 40.0
& 19.7 \\

\textbf{ProSG-B}
& -- & 243 & 7.5 & 30.5 & 125
& \underline{36.7}
& 12.1 \\

\textbf{ProSG-B$^{*}$}
& -- & 243 & 7.5 & 30.5 & 125
& 36.9
& 12.1 \\

\textbf{ProSG-L}
& -- & 243 & 12.2 & 49.5 & 203
& \textbf{36.0}
& \textbf{11.4} \\
\addlinespace[2pt]
\multicolumn{3}{r}{\itshape ProSG-L vs. SAMA-L(best prior)}
& \textcolor{gainGreen}{\scriptsize $\downarrow 29.48\%$}
& \textcolor{gainGreen}{\scriptsize $\downarrow 6.95\%$}
& \textcolor{gainGreen}{\scriptsize $\downarrow 6.88\%$}
& \textcolor{gainGreen}{\scriptsize $\downarrow 2.44\%$}
& \textcolor{gainGreen}{\scriptsize $\downarrow 4.20\%$} \\

\bottomrule
\end{tabular}
}
\end{table*}

\subsection{Implementation Details}
\textbf{Experimental Settings.} All experiments are implemented in PyTorch on a single RTX 5090 GPU, based on the MotionAGFormer
codebase~\cite{mehraban2024motionagformer}.
Following prior works~\cite{tang2023ftcm,tang20233d,zhu2023motionbert},
we apply horizontal flipping during training and testing.
For Human3.6M, we use 2D keypoints with confidence scores
detected by the Stacked Hourglass (SH) ~\cite{newell2016stacked}, following ~\cite{zhu2023motionbert,mehraban2024motionagformer}.
Ground-truth 2D poses are additionally used to evaluate
the model's upper-bound performance. For MPI-INF-3DHP, we use ground-truth 2D poses of 17 joints and the evaluation following the protocol in~\cite{zhao2023poseformerv2,tang20233d}.
We train the network for 90 epochs using AdamW~\cite{adamW}
with a batch size of 4 and a weight decay of 0.01.
The initial learning rate is $5\times10^{-4}$ and is
exponentially decayed by a factor of 0.99 per epoch. Our main and ablation results are averaged over three random
seeds. Detailed standard deviations are reported in the
supplementary material.

\textbf{Model Variants.} We design three variants of our proposed method with different stages to balance accuracy and efficiency. Each variant consists of multiple stacked stages. The base model (ProSG-B) contains $L=16$ stages with a hidden dimension of $C=128$. The other variants are summarized in Supplementary Material.

\begin{table*}[h]
\centering
\caption{Detailed comparison on Human3.6M across actions. Lower errors are better. Best and second-best distinct values among all methods are shown in \textbf{bold} and \underline{underlined}, respectively; ties receive the same formatting. }
\label{tab:comparison}
\setlength{\tabcolsep}{2pt}
\resizebox{\textwidth}{!}{
\begin{tabular}{l|l|c| *{16}{c}}
\toprule
\textbf{Protocol \#1(MPJPE)} & Year & $T$ & Dire. & Disc. & Eat & Greet & Phone & Photo & Pose & Purch. & Sit & SitD & Smoke & Wait & WalkD & Walk & WalkT & Avg \\
\midrule
MixSTE~\cite{zhang2022mixste} & 2022 & 243
& 37.6 & 40.9 & 37.3 & 39.7 & 42.3 & 49.9 & 40.1 & 39.8
& 51.7 & 55.0 & 42.1 & 39.8 & 41.0 & 27.9 & 27.9 & 40.9 \\
STCFormer~\cite{tang20233d} & 2023 & 243
& 39.6 & 41.6 & 37.4 & 38.9 & 43.1 & 51.1 & 39.1 & 39.7
& 51.4 & 57.4 & 41.8 & 38.5 & 40.7 & 27.1 & 28.6 & 41.0 \\
MotionBERT(scratch)~\cite{zhu2023motionbert} & 2023 & 243
& 36.6 & 39.3 & 37.8 & 33.5 & 41.4 & 49.9 & 37.0 & 35.5
& 50.4 & 56.5 & 41.4 & 38.2 & 37.3 & 26.2 & 26.9 & 39.2 \\
MotionAGFormer~\cite{mehraban2024motionagformer} & 2024 & 243
& 36.8 & 38.5 & 35.9 & 33.0 & 41.1 & 48.6 & 38.0 & 34.8
& 49.0 & \underline{51.4} & 40.3 & 37.4 & 36.3 & 27.2 & 27.2 & 38.4 \\
KTPFormer~\cite{peng2024ktpformer} & 2024 & 243
& 37.3 & 39.2 & 35.9 & 37.6 & 42.5 & 48.2 & 38.6 & 39.0
& 51.4 & 55.9 & 41.6 & 39.0 & 40.0 & 27.0 & 27.4 & 40.1 \\
SCTFormer~\cite{zheng2025spectral} & 2026 & 243
& 35.8 & 38.3 & \underline{35.7} & 32.2 & \underline{39.7} & 47.3 & 36.0 & 34.5
& 50.0 & 52.9 & 39.4 & 36.8 & 35.7 & 25.2 & 25.3 & 37.7 \\
\rowcolor{gray!20} \textbf{ProSG-S} & -- & 81
& 36.9 & 40.3 & 38.3 & 33.9 & 42.7 & 48.6 & 38.8 & 36.3
& 51.9 & 59.0 & 42.1 & 39.4 & 38.1 & 26.6 & 27.8 & 40.0 \\
\rowcolor{gray!20} \textbf{ProSG-B} & -- & 243
& \underline{34.5} & \underline{37.2} & 35.8 & \underline{31.6} & \underline{39.7} & \underline{43.9} & \underline{35.6} & \underline{33.1}
& \textbf{47.7} & 54.0 & \underline{39.2} & \textbf{34.9} & \textbf{33.7} & \underline{25.0} & \underline{25.1} & \underline{36.7} \\
\rowcolor{gray!20} \textbf{ProSG-L} & -- & 243
& \textbf{34.0} & \textbf{36.0} & \textbf{35.3} & \textbf{30.5} & \textbf{39.4} & \textbf{43.3} & \textbf{34.3} & \textbf{33.0}
& \underline{47.8} & \textbf{50.9} & \textbf{38.5} & \underline{35.3} & \underline{33.8} & \textbf{23.6} & \textbf{23.9} & \textbf{36.0} \\
\midrule
\textbf{Protocol \#2(P-MPJPE)} & Year & $T$ & Dire. & Disc. & Eat & Greet & Phone & Photo & Pose & Purch. & Sit & SitD & Smoke & Wait & WalkD & Walk & WalkT & Avg \\
\midrule
MixSTE~\cite{zhang2022mixste} & 2022 & 243
& 30.8 & 33.1 & 32.3 & 33.7 & 33.1 & 39.1 & 31.1 & 30.5
& 42.5 & \textbf{44.5} & 34.1 & 30.8 & 32.7 & 22.1 & 22.6 & 32.6 \\
P-STMO~\cite{shan2022p} & 2022 & 243
& 35.2 & 35.2 & 32.9 & 33.9 & 35.4 & 39.1 & 32.5 & 31.5
& 44.6 & 48.2 & 36.3 & 32.9 & 34.4 & 23.8 & 23.9 & 35.4 \\
MotionBERT(scratch)~\cite{zhu2023motionbert} & 2023 & 243
& \underline{29.5} & 33.8 & 31.7 & 31.7 & 34.3 & 38.9 & 30.1 & 30.2
& 41.4 & 47.6 & 35.0 & 30.9 & 32.7 & 25.3 & 23.6 & 33.1 \\
MotionAGFormer~\cite{mehraban2024motionagformer} & 2024 & 243
& 31.0 & 32.6 & 31.0 & 27.9 & 34.0 & 38.7 & 31.5 & 30.0
& 41.4 & 45.4 & 34.8 & 30.8 & 31.3 & 22.8 & 23.2 & 32.5 \\
KTPFormer~\cite{peng2024ktpformer} & 2024 & 243
& 30.1 & 32.3 & \textbf{29.6} & 30.8 & \textbf{32.3} & 37.3 & 30.0 & 30.2
& 41.0 & \underline{45.3} & \textbf{33.6} & 29.9 & 31.4 & 21.5 & 22.6 & 31.9 \\
SCTFormer~\cite{zheng2025spectral} & 2026 & 243
& 29.7 & 31.6 & 30.6 & 27.2 & 33.1 & 37.6 & 29.6 & 29.9
& 41.6 & 46.6 & 34.5 & 29.9 & 30.6 & 21.4 & \underline{21.8} & 31.7 \\
\rowcolor{gray!20} \textbf{ProSG-S} & -- & 81
& 31.2 & 33.6 & 33.0 & 28.6 & 35.2 & 39.9 & 31.9 & 31.3
& 43.9 & 51.0 & 36.2 & 32.2 & 32.9 & 22.5 & 24.1 & 33.8 \\
\rowcolor{gray!20} \textbf{ProSG-B} & -- & 243
& \textbf{29.4} & \underline{31.4} & 30.4 & \underline{26.7} & \underline{32.8} & \underline{36.0} & \underline{29.3} & \underline{28.6}
& \underline{40.8} & 46.8 & 34.2 & \textbf{28.7} & \underline{29.2} & \underline{21.1} & \underline{21.8} & \underline{31.1} \\
\rowcolor{gray!20} \textbf{ProSG-L} & -- & 243
& \textbf{29.4} & \textbf{30.6} & \underline{30.0} & \textbf{25.8} & 33.0 & \textbf{35.5} & \textbf{28.1} & \textbf{28.5}
& \textbf{40.7} & \textbf{44.5} & \underline{33.9} & \underline{28.8} & \textbf{29.1} & \textbf{20.1} & \textbf{20.9} & \textbf{30.6} \\
\midrule
\textbf{MPJPE(GT)} & Year & $T$ & Dire. & Disc. & Eat & Greet & Phone & Photo & Pose & Purch. & Sit & SitD & Smoke & Wait & WalkD & Walk & WalkT & Avg \\
\hline
MixSTE~\cite{zhang2022mixste} & 2022 & 243
& 21.6 & 22.0 & 20.4 & 21.0 & 20.8 & 24.3 & 24.7 & 21.9
& 26.9 & 24.9 & 21.2 & 21.5 & 20.8 & 14.7 & 15.7 & 21.6 \\
STCFormer~\cite{tang20233d} & 2023 & 243
& 20.8 & 21.8 & 20.0 & 20.6 & 23.4 & 25.0 & 23.6 & 19.3
& 27.8 & 26.1 & 21.6 & 20.6 & 19.5 & 14.3 & 15.1 & 21.3 \\
MotionBERT(scratch)~\cite{zhu2023motionbert} & 2023 & 243 & 16.7 & 19.9 & 17.1 & 16.5 & 17.4 & 18.8 & 19.3 & 20.5 & 24.0 & 22.1 & 18.6 & 16.8 & 16.7 & 10.8 & 11.5 & 17.9 \\
KTPFormer~\cite{peng2024ktpformer} & 2024 & 243
& 19.6 & 18.6 & 18.5 & 18.1 & 18.7 & 22.1 & 20.8 & 18.3
& 22.8 & 22.4 & 18.8 & 18.1 & 18.4 & 13.9 & 15.2 & 19.0 \\
SCTFormer~\cite{zheng2025spectral} & 2026 & 243
& 14.1 & 15.8 & 16.1 & 14.4 & 15.8 & 16.6 & 16.4 & 16.6
& 20.9 & 20.6 & 16.0 & 14.5 & 14.9 & 9.0 & 9.8 & 15.4 \\
\rowcolor{gray!20}\textbf{ProSG-S} & -- & 81
& 18.6 & 19.6 & 21.0 & 17.3 & 20.2 & 22.9 & 20.7 & 20.6
& 28.4 & 30.1 & 21.5 & 18.7 & 19.7 & 15.6 & 16.1 & 19.7 \\
\rowcolor{gray!20}\textbf{ProSG-B} & -- & 243
& \underline{11.7} & \underline{12.1} & \underline{13.4} & \underline{10.9} & \underline{12.3} & \underline{13.2} & \underline{12.1} & \underline{13.5}
& \underline{17.7} & \underline{16.5} & \underline{12.7} & \underline{10.7} & \underline{10.7} & \underline{7.2} & \underline{7.3} & \underline{12.1} \\
\rowcolor{gray!20}\textbf{ProSG-L} & -- & 243
& \textbf{10.8} & \textbf{11.7} & \textbf{12.6} & \textbf{10.4} & \textbf{11.9} & \textbf{12.7} & \textbf{11.2} & \textbf{11.7}
& \textbf{17.5} & \textbf{15.7} & \textbf{12.2} & \textbf{10.6} & \textbf{10.2} & \textbf{6.1} & \textbf{6.6} & \textbf{11.4} \\
\bottomrule
\end{tabular}}
\end{table*}

\begin{table}[h]
\centering
\caption{Quantitative Results on MPI-INF-3DHP. $T$ denotes the number of input frames in the 3DHP dataset. }
\small
\label{tab:mpi}
\begin{tabular}{l|c|c|c|c}
\toprule
Method & $T$ & PCK$\uparrow$ & AUC$\uparrow$ & MPJPE$\downarrow$ \\
\midrule
Anatomy3D~\cite{chen2021anatomy} 
& 81 & 87.8 & 53.8 & 79.1\\
P-STMO~\cite{shan2022p}
& 81 & 97.9 & 75.8 & 32.2 \\
STCFormer~\cite{tang20233d}
& 81 & 98.7 & 83.9 & 23.1 \\
PoseFormerV2~\cite{zhao2023poseformerv2}
& 81 & 97.9 & 78.8 & 27.8 \\
GLA-GCN~\cite{yu2023gla}
& 81 & 98.5 & 79.1 & 27.7 \\
MotionAGFormer-B~\cite{mehraban2024motionagformer}
& 81 & 98.3 & 84.2 & 18.2 \\
SCTFormer~\cite{zheng2025spectral}
& 81 & 98.7 & 86.6 & 16.0 \\
PoseMamba~\cite{huang2025posemamba}
& 81 & -- & -- & 14.5 \\
PoseMagic~\cite{zhang2025pose}
& 81 & 98.8 & 87.6 & 14.7 \\
SAMA~\cite{lu2025structure}
& 81 & 99.0 & 88.3 & 14.4 \\
\midrule
\textbf{ProSG-S}
& 81 & \textbf{99.1} & 89.0 & 14.3 \\
\textbf{ProSG-B}
& 81 & 98.9 & \textbf{89.2} & \underline{13.4} \\
\textbf{ProSG-L}
& 81 & \underline{99.0} & \textbf{89.2} & \textbf{13.1} \\
\addlinespace[2pt]
\multicolumn{2}{l}{\itshape ProSG-L vs. SAMA(best prior)}
& \scriptsize --
& \textcolor{gainGreen}{\scriptsize $\uparrow 1.02\%$}
& \textcolor{gainGreen}{\scriptsize $\downarrow 9.03\%$} \\
\bottomrule
\end{tabular}
\end{table}

\textbf{Analysis on Human3.6M.}
Tables~\ref{tab:computational-cost}
and~\ref{tab:comparison} show that ProSG-L achieves
$36.0$ mm MPJPE and $30.6$ mm P-MPJPE,
outperforming Transformer-based methods.
It surpasses Transformer-based baselines
MotionAGFormer~\cite{mehraban2024motionagformer}
and TCPFormer~\cite{liu2025tcpformer} in accuracy,
with lower computational cost and fewer parameters than TCPFormer.
Against the Mamba-based baseline SAMA-L~\cite{lu2025structure},
ProSG-L achieves lower MPJPE with fewer parameters and MACs.
Per-action results show competitive performance under both
protocols, particularly on \textit{Walking}
and \textit{Walking Together}.

\textbf{Analysis on MPI-INF-3DHP.} 
We further evaluate our method on MPI-INF-3DHP.
Following prior works~\cite{mehta2017monocular, zhang2022mixste},
the model is trained using ground-truth 2D poses as inputs.
As shown in Table~\ref{tab:mpi}, ProSG-L achieves an AUC of
89.2 and an MPJPE of 13.1 mm, improving over SAMA and
PoseMagic while matching SAMA's PCK of 99.0. Specifically, ProSG-L reduces MPJPE by 1.3 mm (9.0\%)
compared with SAMA and by 1.6 mm (10.9\%) compared with PoseMagic.
These results demonstrate the effectiveness of our framework
on this benchmark.

\subsection{Ablation studies}
\begin{table}[h]
\centering
\caption{Ablation of SGA and SCR on Human3.6M.}
\label{tab:components}
\resizebox{\linewidth}{!}{
\begin{tabular}{c|c c c |c c c c}
\toprule
Step & Backbone & SGA & SCR & MPJPE $\downarrow$ & P-MPJPE $\downarrow$ & Params (M)$\downarrow$ & MACs (G)$\downarrow$\\
\hline
1 & \checkmark & - & - &  38.4 & 32.1 & 6.4 & 30.8 \\
2 & \checkmark & \checkmark & - &  37.1 & 31.5 & 6.4 & 31.0 \\
3 & \checkmark & - & \checkmark &  38.2 & 31.9 & 7.5 & 30.3 \\
ProSG & \checkmark & \checkmark & \checkmark &  \textbf{36.7} & \textbf{31.1} & 7.5 & 30.5 \\
\bottomrule
\end{tabular}}
\vspace{-0.5em}
\end{table}

\begin{table}[t]
\centering
\caption{Comparison with sublayer residual aggregation and
ablation of the internal components of SGA.}
\label{tab:sga_ablation}
\setlength{\tabcolsep}{4pt}
\renewcommand{\arraystretch}{1.12}
\resizebox{0.95\columnwidth}{!}{
\begin{tabular}{lcc}
\toprule
Configuration & MPJPE $\downarrow$ & P-MPJPE $\downarrow$ \\
\midrule
Backbone
& 38.4 & 32.1 \\

Block AttnRes (Attention/MLP sublayers grouped)
& 38.3 & 32.0 \\

Full AttnRes (Attention/MLP sublayers)
& 38.1 & 31.9 \\

Cross-stage Structural Aggregation
& 37.4 & 31.6 \\

+ Structural Retrieval Scoring (SGA)
& \textbf{37.1} & \textbf{31.5} \\
\bottomrule
\end{tabular}
}
\vspace{-0.5em}
\end{table}

\begin{table}[t]
\centering
\caption{Analysis of cross-stage representation similarity (CKA)
and  performance under different architectures.
``w/o CT'' denotes SCR without compact token operation. STCBlock is trained and evaluated under the same protocol as our backbone.}
\label{tab:cka_analysis}
\setlength{\tabcolsep}{4pt}
\renewcommand{\arraystretch}{1.12}
\resizebox{0.9\columnwidth}{!}{
\begin{tabular}{llcc}
\toprule
\textbf{Architecture} & \textbf{Configuration}
& \textbf{Mean CKA} $\uparrow$
& \textbf{MPJPE} $\downarrow$ \\
\midrule
Sequential & Backbone       & 0.14 & 38.4  \\
\midrule
Hybrid     & Backbone       & 0.24 & 37.5  \\
\midrule
\multirow{6}{*}{Parallel}
 & STCBlock (our reproduction) & 0.29 & 38.8 \\
 & backbone & 0.31 & 38.4 \\
 & + SGA & 0.32 & 37.1 \\
 & + SGA + Standard ST-Attention & 0.32 & 37.4 \\
 & + SGA + SCR (w/o CT) & 0.30 & 37.2 \\
 & + SGA + SCR & \textbf{0.38} & \textbf{36.7} \\
\bottomrule
\end{tabular}
}
\vspace{-1.5em}
\end{table}


\textbf{Impact of Each Component.}
We conduct ablation studies to evaluate the effectiveness of SGA and SCR, as shown in Table~\ref{tab:components}. Incorporating SGA reduces MPJPE from 38.4 mm to 37.1 mm. Notably, SCR is designed to complement SGA by reorganizing intermediate pose representations for more effective structural feature reuse. Their combination further reduces MPJPE to 36.7 mm and P-MPJPE to 31.1 mm, highlighting the complementary roles of representation compatibility and adaptive cross-stage aggregation.

\begin{figure*}[!t]
  \centering
   \includegraphics[width=0.86\linewidth]{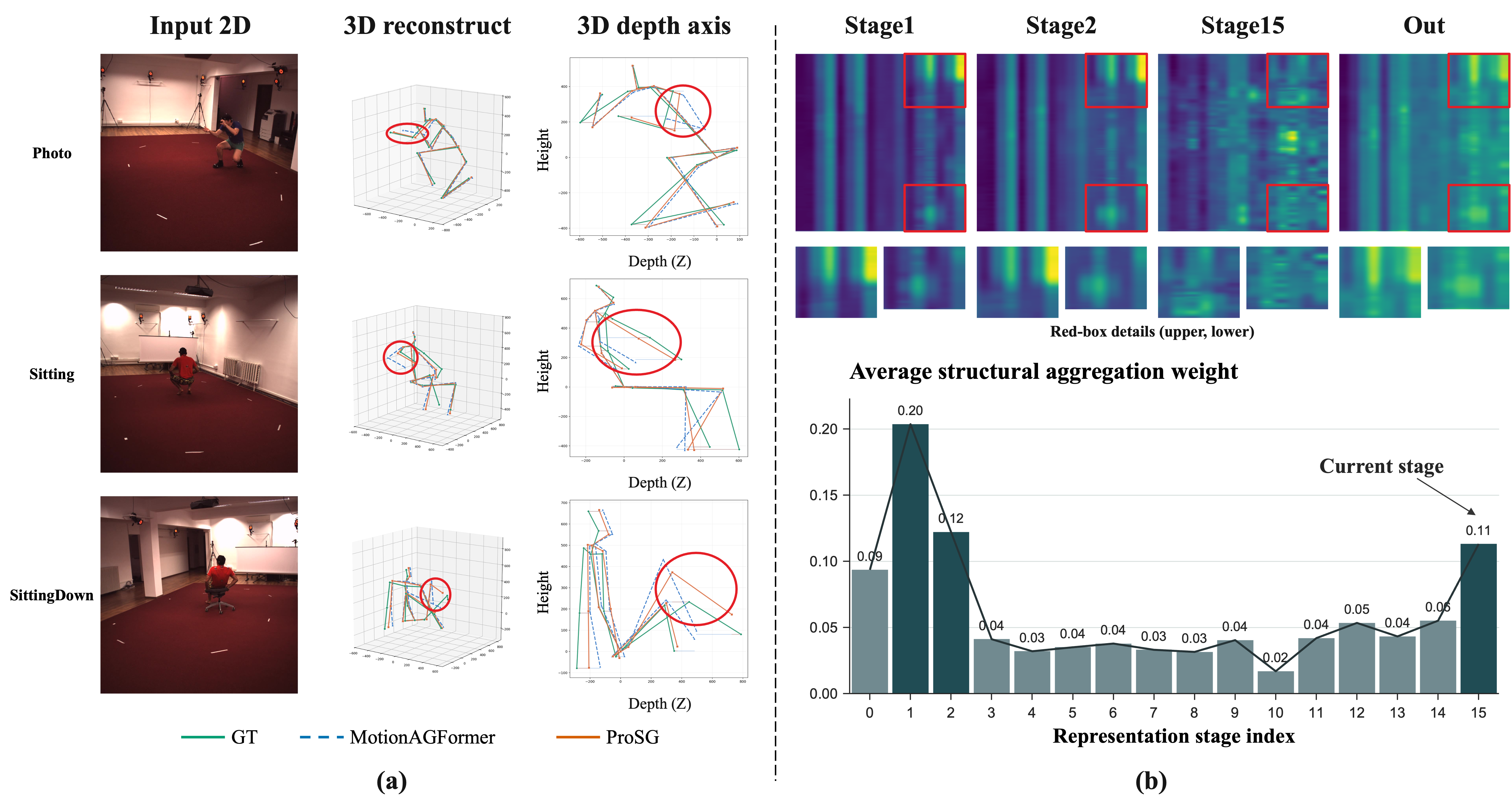}
\vspace{-1em}
   \caption{(a): comparison results on Human3.6M. (b): Stage-wise structural retrieval weights at the final stage, together with feature responses from selected stages and the aggregated output.
Additional examples are provided in the supplementary material. }
    \label{fig:vis}
\end{figure*}

\begin{figure}[h]
  \centering
   \includegraphics[width=0.90\linewidth]{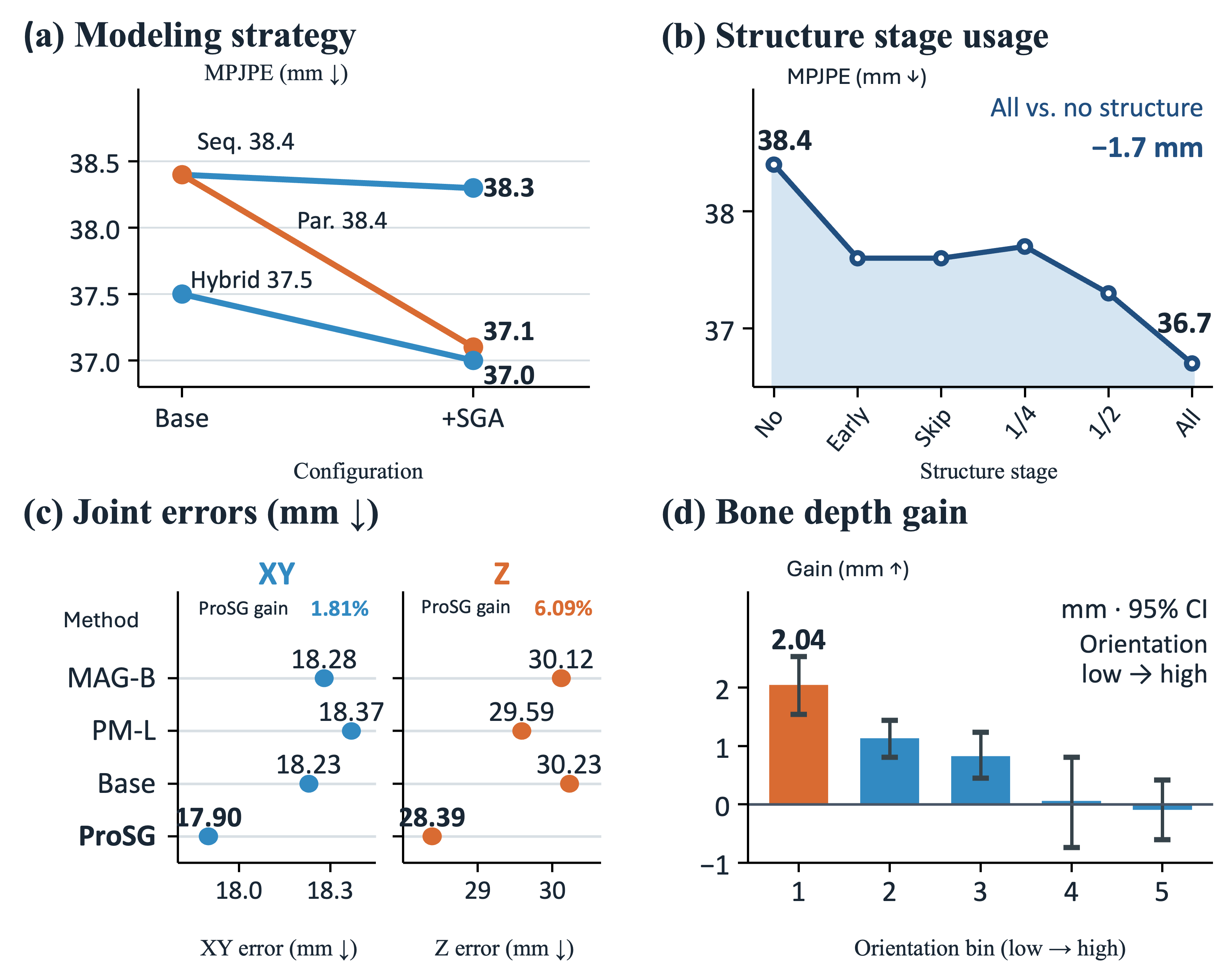}

   \caption{(a) SGA gains under sequential, parallel, and hybrid
architectures. (b) Performance with different structural representation selection strategies (c) Joint errors in the image plane (XY) and along the camera-space depth axis (Z). PM-L and MAG-B denote PoseMamba~\cite{huang2025posemamba} and MotionAGFormer~\cite{mehraban2024motionagformer}. (d) Reduction in bone-relative depth error (backbone minus ProSG) across five groups ordered by increasing $|\Delta Z|/\text{bone length}$. Lower groups correspond to bones more parallel to the image plane.}
   \label{fig:Effect of History}
   \vspace{-1em}
\end{figure}

\textbf{Effect of Structural Information Guidance.}
As shown in Table~\ref{tab:sga_ablation}, Attention
Residual~\cite{team2026attention} aggregates features produced by the attention and MLP sublayers. 
Block AttnRes yields 38.3 mm MPJPE on our backbone,
while even Full AttnRes reduces the error only to 38.1 mm.
In contrast, organizing complete stage outputs as structural representations for structure aggregation reduces MPJPE to 37.4 mm. Further incorporating skeletal-neighbor context and temporal score smoothing forms the complete SGA and reduces MPJPE to 37.1 mm. Fig.~\ref{fig:Effect of History}(a) further shows that the parallel architecture benefits more from SGA. Fig.~\ref{fig:Effect of History}(b) compares different stage-selection strategies, including varying the number of stages used, and skipping stages, with full-stage aggregation achieving the best accuracy.

\textbf{Effect of Structural Context Reorganization.}
For comparison, we replace SCR with standard spatio-temporal
attention at the same position within each stage, retaining SGA. As shown in Table~\ref{tab:cka_analysis},
SCR outperforms standard spatio-temporal attention
(36.7 vs. 37.4 mm), while removing compact contextual
tokens increases the error to 37.2 mm.
Compared with SGA alone, SCR increases CKA
from 0.32 to 0.38 while reducing MPJPE from
37.1 to 36.7 mm. These results demonstrate the
effectiveness of contextual reorganization for
structural feature guidance, accompanied by
increased inter-stage representation similarity.
We set the numbers of contextual representatives to
$M_s=9$ and $M_t=27$, sensitivity analyses of the representative counts $M_t$ and
$M_s$ are provided in the supplementary material.

\textbf{Axis-wise Error Analysis.}
We evaluate errors in root-relative camera coordinates.
XY error is the mean Euclidean distance between predicted
and ground-truth joint positions in the image plane,
whereas Z error is the mean absolute difference along
the camera depth axis. As shown in Fig.~\ref{fig:Effect of History}(c), ProSG achieves lower error than existing methods~\cite{huang2025posemamba,mehraban2024motionagformer}, along the camera-space
depth axis. This suggests
that structural cues provide effective guidance for subsequent depth inference. Fig.~\ref{fig:Effect of History}(d) compares bone-relative depth errors across five  groups,  ProSG achieves the largest reduction over the backbone in Group 1. This suggests that reusing structural cues helps
constrain relative joint depths for bones extending
along the image plane, reducing erroneous
out-of-plane inclination during 3D reconstruction.

\textbf{Qualitative Analysis.}
Fig.~\ref{fig:vis}(a) compares reconstructed poses and highlights differences along the depth axis. ProSG estimates depth more accurately than MotionAGFormer for bones nearly parallel to the image plane. Fig.~\ref{fig:vis}(b) shows stage-wise structural retrieval weights and feature responses. Early stages receive substantial weights, and the highlighted output regions retain patterns visible in early-stage features, consistent with the reuse
of earlier representations during reconstruction.

%% file: sec/conclusion.tex
\section{Conclusion}
In this paper, we investigate the complementary role of structural information encoded at different stages of 3D human pose estimation. We propose ProSG, a progressive structural guidance framework that adaptively integrates pose representations through Structure Guided Accumulation and Structural Context Reorganization, enabling early structural cues and intermediate representations to support subsequent 3D inference. Extensive experiments demonstrate that ProSG achieves state-of-the-art performance with favorable computational efficiency.